\documentclass[final]{cvpr}

\usepackage{times}
\usepackage{epsfig}
\usepackage{graphicx}
\usepackage{amsmath}
\usepackage{amssymb}
\usepackage{multicol}
\usepackage[title]{appendix}

\newcommand{\method}{BARS\xspace}

\usepackage{color}
\usepackage{xcolor}
\usepackage{xspace}

\usepackage{multirow}
\usepackage{booktabs}
\usepackage{amsmath}
\newcommand{\Sign}{\mathrm{Sign}}

\DeclareMathOperator*{\argmin}{arg\,min}

\usepackage[pagebackref=true,breaklinks=true,colorlinks,bookmarks=false]{hyperref}

\begin{document}

\title{\method: Joint Search of Cell Topology and Layout for Accurate and Efficient Binary ARchitectures}

\author{
  Tianchen Zhao$^{123*}$\\
  {\tt\small ztc16@buaa.edu.cn}
\and
Xuefei Ning$^{12*}$\\
{\tt\small foxdoraame@gmail.com}
\and
Xiangsheng Shi$^{12*}$\\
{\tt\small shi-xs20@mails.tsinghua.edu.cn}
\and 
Songyi Yang$^{2}$\\
{\tt\small songyi-yang@outlook.com}
\and
Shuang Liang$^{12}$\\
{\tt\small shuang.liang@novauto.com.cn}
\and
Peng Lei$^{3}$\\
{\tt\small buaaray@gmail.com}
\and
Jianfei Chen$^{4}$\\
{\tt\small chrisjianfeichen@gmail.com}
\and
Huazhong Yang$^{1\dagger}$\\
{\tt\small yanghz@tsinghua.edu.cn}
\and
Yu Wang$^{1\dagger}$\\
{\tt\small yu-wang@tsinghua.edu.cn}}


\twocolumn[
\begin{@twocolumnfalse}
\maketitle
\begin{center}
{\normalsize 
\vspace{-15pt}
$^1$Department of Electronic Engineering, Tsinghua University\quad $^2$Novauto\\
$^3$Department of Electronic Engineering, Beihang University\\
$^4$Department of Computer Science, Tsinghua University\\}
\end{center}
\end{@twocolumnfalse}
]
{
  \renewcommand{\thefootnote}%
    {\fnsymbol{footnote}}
    \footnotetext[1]{Equal contribution. Work done while T. Zhao is visiting Tsinghua University.}
      \renewcommand{\thefootnote}%
    {\fnsymbol{footnote}}
  \footnotetext[2]{Corresponding authors. }
}

%
\begin{abstract}

Binary Neural Networks (BNNs) have received significant attention due to their promising efficiency. Currently, most BNN studies directly adopt widely-used CNN architectures, which can be suboptimal for BNNs. This paper proposes a novel Binary ARchitecture Search (BARS) flow to discover superior binary architecture in a large design space. Specifically, we analyze the information bottlenecks that are related to both the topology and layout architecture design choices. And we propose to automatically search for the optimal information flow. To achieve that, we design a two-level (Macro \& Micro) search space tailored for BNNs, and apply differentiable neural architecture search (NAS) to explore this search space efficiently. The macro-level search space includes width and depth decisions, which is required for better balancing the model performance and complexity. We also design the micro-level search space to strengthen the information flow for BNN. 
On CIFAR-10, \method achieves $1.5\%$ higher accuracy with $2/3$ binary operations and $1/10$ floating-point operations comparing with existing BNN NAS studies. On ImageNet, with similar resource consumption, \method-discovered architecture achieves a $6\%$ accuracy gain than hand-crafted binary ResNet-18 architectures, and outperforms other binary architectures while fully binarizing the architecture backbone.


\end{abstract}

\section{Introduction}
\label{sec:intro}

Convolutional Neural Networks (CNNs) have demonstrated great performance in computer vision tasks. However, CNNs often require substantial computational and storage resources, making their deployment difficult for edge devices and other resource-constrained scenarios. Existing approaches to alleviate this problem include network pruning~\cite{deepcompression,grouplasso}, efficient architecture design~\cite{mobv2,shufflenet,mnasnet}, and quantization~\cite{hubara2017quantizednn}. Among them, Binary Neural Network (BNN) is a promising direction that utilizes 1-bit quantization. By binarizing network parameters and activations, the resource-hungry 32-bit floating-point multiplications can be replaced by efficient bitwise operations (e.g. XNOR, bitcount), significantly reducing the computation and memory burden. Despite their computational efficiency, BNNs often suffer from unsatisfactory  performance through the binarization process. Moreover, parts of the network computation flow still remain in full-precision (FP), bringing difficulty for the hardware acceleration.

\begin{figure}[t]
    \centering
    \includegraphics[width=1.0\linewidth]{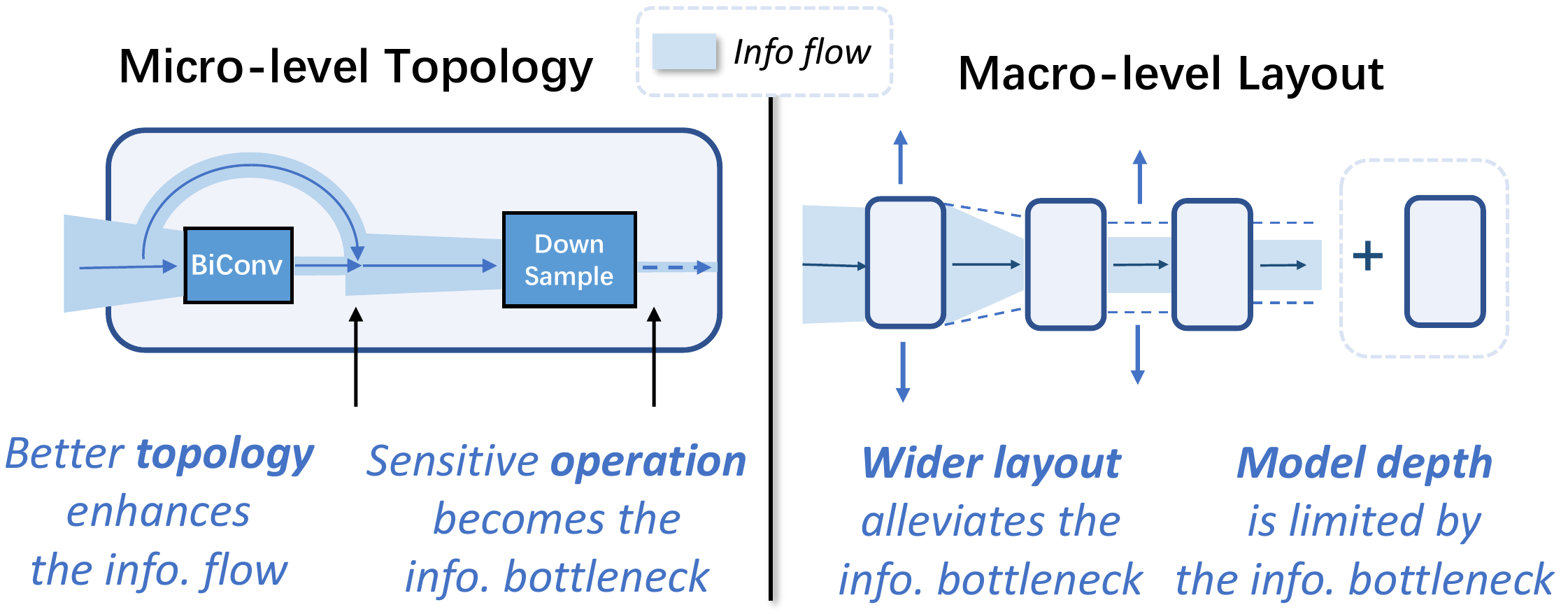}
    \caption{\textbf{The information bottleneck phenomenon related to micro-level topology and macro-level layout.}} 
    \label{fig:info_bot}
\end{figure}


Architectural design is critical for BNNs' performance. However, existing architectures for FP networks are suboptimal for BNNs. 
As BNNs' activations are binary, they carry much less information relative to their FP counterparts, causing the \emph{information bottleneck}. Enhancing the information flow in BNNs is thus critical for their performance~\cite{Bethge2018TrainingCB}.
Viewing the network architecture as a sequence of basic building blocks (or cells), the information bottleneck affects BNNs' performance through two levels of granularity: 
(1) The \emph{micro-level} considers the inner structure of each cell (i.e. cell topology) . At the micro-level, compact CNN blocks (e.g. depthwise, bottleneck convs) and scale-altering layers (e.g. downsampling) are the information bottlenecks that hinder the performance of BNNs. On the other hand, shortcut connections strengthen the information flow~\cite{mobinet,bethge2019binarydensenet}.
(2) The \emph{macro-level} focuses on the composition structure of cells (i.e. cell layout). At the macro-level, expanding networks' width allows activations to carry more information, which could strengthen the information flow and bring performance gain~\cite{widen_and_squeeze}. In contrast, deeper architectures might not work well, since adding more layers does not alleviate existing information bottlenecks. 
Fig.~\ref{fig:info_bot} illustrates these design considerations under the information flow perspective. 

Neural architecture search (NAS)~\cite{enas} is a promising direction to find  optimal architectures for neural networks automatically. However, existing NAS studies for BNNs~\cite{bnas-width, bnas} mostly leverage the search framework for FP networks~\cite{darts}, and did not fully consider preserving the information flow for both the micro and macro-level. For example, 
\cite{bnas,bats} employ micro-level topology search, but use a pre-defined cell layout and model augmentation scheme identical to FP networks. \cite{bnas-width} searches for the macro-level layout through determining layer-wise widths, but it uses a fixed topology adapted from CNNs. 

In this paper, we propose \method, a BNN-oriented differentiable NAS flow, in which the search space, along with the search and derive strategies are carefully designed and developed according to the characteristics of BNNs. Unlike existing works~\cite{bnas-width, bats, bnas}, \method extends the original micro-level DARTS~\cite{darts} search space to the macro-level. And jointly searches for the micro-level cell topology and the macro-level cell layout with a 2-level search space. 
We design a novel macro-level depth \& width search space that could be unified in the differentiable NAS framework. It seeks to strike a better balance between model performance and complexity. We also improve the micro-level search space for automatically discovering topologies that avoid creating bottlenecks and maintain proper information flow. Besides, we propose improvements on the search strategy such as Gumbel sampling and entropy regularization to ensure a stabilized search in a much bigger search space.

With the above mentioned techniques, BARS-discovered architecture outperforms CNN-adapted binary architectures by a large margin ($6\%$ better accuracy than hand-crafted binary ResNet18 on ImageNet).
It also achieves superior performance than state-of-the-art baseline architectures with smaller complexity. Furthermore, BARS reduces the full-precision operations significantly. BARS-discovered architectures only have  10\% floating-point operations compared with existing BNN NAS studies on CIFAR.

\begin{figure*}[t]
    \centering
    \includegraphics[width=1.0\linewidth]{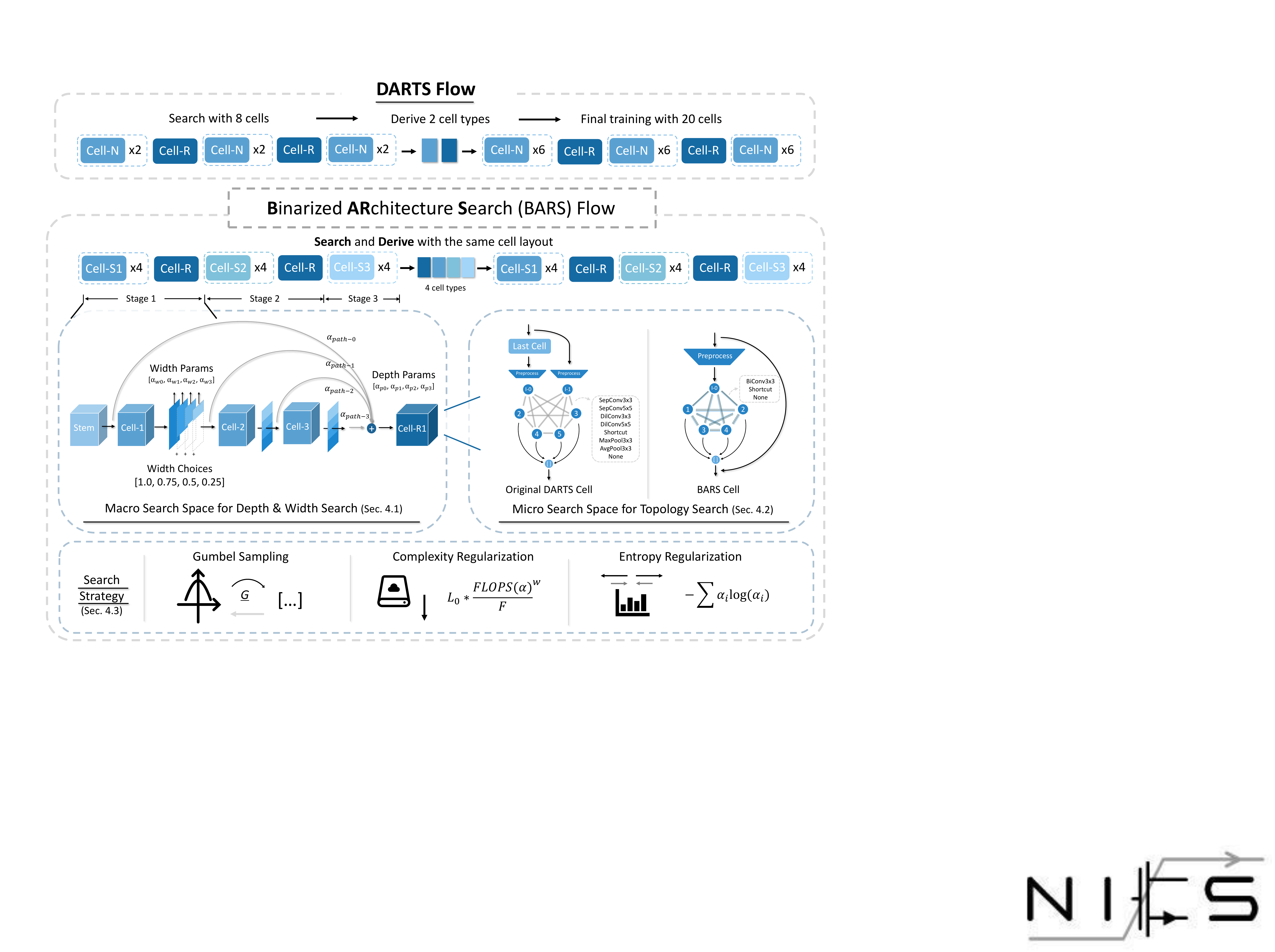}
    \caption{\textbf{The workflow of our proposed Binary ARchitecture Search (\method) (lower) vs. DARTS (upper)~\cite{darts,bats,BinarizedNAS}}. 
    \method tailors a series of modifications on search space and search strategies to facilitate a proper search process for BNNs.} 
    \label{fig:main-flow}
\end{figure*}

\section{Related Works}
\label{sec:rw}

\subsection{Binary Neural Networks}

\noindent\textbf{Binarization Scheme}
Network binarization could be viewed as an extreme case of network quantization. It could replace the original FP32 multiplications with efficient bitwise operations, gaining over 10 times the processing speed. 
However, due to the lack of representation ability, binary neural networks often suffer from noticeable accuracy degradation.
Several methods have been proposed to improve the performance of BNN. XNORNet\cite{xnornet} uses shared scaling factors to improve the representation ability without introducing much computational overhead. Many recent studies~\cite{reactnet,bats,bnas} follow its binarization scheme, and so do we. 
Some other binarization schemes are also proposed, such as: \cite{xnornet++} fuses the weight and activation scaling factor together before inference. 
Other approaches for improving BNN performance focus on minimizing the quantization error~\cite{lqnet}, redesigning the training loss~\cite{apprentice}, or amending the gradient estimation~\cite{bireal, irnet}.

\noindent\textbf{Binary Architectural Advances}
The aforementioned methods mainly focus on improving the binarization or training scheme. Furthermore, the network architecture also plays a critical role in determining the performance of a BNN. Previous studies mainly address the information bottleneck issue from 2 perspectives: 1) Strengthen the information flow by adding more shortcuts~\cite{bireal,Bethge2018TrainingCB,bethge2019binarydensenet}. 2) Identify and eliminate some information bottleneck manually: Most of the recent studies~\cite{bireal,irnet} adopt full-precision downsampling layer; \cite{mobinet} modifies separable convolutions in the MobileNet architectures. 

\subsection{Neural Architecture Search (NAS)}

\noindent\textbf{NAS Search Space }
NAS search space designs in recent studies can be divided into two categories: macro-level and micro-level (cell-level). The macro-level describes how cells are organized to construct the entire architecture, and methods have been developed to search for these layout decisions, including width and depth~\cite{fbnet,tfnas}. On the other hand, the micro-level describes the connecting operations inside each cell, and aims to find a superior intra-cell topology. There exist many studies that only search for the micro-level cell topology and organize cells into a pre-defined layout~\cite{darts}. In this paper, we employ both the macro- and micro-level search to obtain accurate and efficient binary architectures.

\noindent\textbf{Differentiable NAS}
Considerable efforts have been devoted to developing and applying gradient-based NAS (i.e. Differentiable NAS) methods~\cite{darts,snas,tfnas} due to its high search efficiency. DARTS~\cite{darts} first models the NAS problem as a bilevel optimization problem, in which the architecture parameters are updated using gradient methods.

Cell-based search spaces~\cite{learningta} are designed to facilitate a more efficient NAS process and have been widely adopted~\cite{darts,snas}. Usually, there are two types of cells in a cell-based search space: normal cells and reduce cells (stride $> 1$). These two types of cells are stacked in a pre-defined order to construct a complete architecture. 

\noindent\textbf{NAS for Binary Architecture}
Previous studies on improving BNN architecture design often adopt minor modifications to existing well-performing CNN models. Applying NAS to the binary domain could be an effective solution to discover more suitable architecture for BNN. ~\cite{bnas-width} strikes a balance between accuracy and resource consumption for BNN via evolutionary search on the network width. ~\cite{Phan2020BinarizingMV} adopts a similar approach on groups in convolutions. ~\cite{bnas,bats} introduce gradient-based search for BNN. They observe that traditional gradient-based NAS methods such as~\cite{darts,snas} can not be directly applied for BNN search. Thus, they modify the operations in search space and the search strategy. These methods make advances in searching for efficient binary architectures. However, they still have the following drawbacks. Firstly, they solely focus on only one of the two closely related aspects: network topology and complexity. Secondly, full-precision layers still exist in the main body of the architecture. 
(~\cite{bats} uses a full-precision preprocess layer in each cell, and ~\cite{bnas} uses a full-precision shortcut for reduction cells, which takes up the majority of the computation). 
BARS aims to search for both the topology and complexity of the binary architecture, as well as pursuing full binarization of the main body of the architecture.

\section{Preliminary}

\subsection{Network Binarization}

\method follows the binarization scheme proposed in XNORNet~\cite{xnornet} with the modification of using a single scaling factor instead of channel-wise for more efficiency. 
The binary convolution of weights $W \in \mathbb{R}^{C_{out}\times C_{in} \times K_w \times K_h}$ and the input feature map $ X \in \mathbb{R}^{bs \times C_{in} \times W \times H}$ can be written as in Eq.~\ref{eq:bnn_conv}, where $C_{out}$ and $C_{in}$ represent the input and output channels respectively. ($K_w$, $K_h$) and ($W$, $H$) are the dimensions of the convolution kernel and of the feature map, and $bs$ is the batch size. 
\begin{equation}
\begin{split}
    W * X = ( \Sign(W) \odot \Sign(X) ) \otimes \beta
\end{split}
\label{eq:bnn_conv}
\end{equation}

Here $\odot$ denotes binary multiplication, which could be simplified into XNOR and bitcount operations. $\otimes$ denotes full precision element-wise multiplication. $\beta$ is a real-valued scaling factor. During inference, the binarization takes place before the convolution. During training, the gradient of the non-differentiable binarization ($\Sign(w)$) is acquired with the Straight-Through~\cite{BinarizedNN} scheme to update the real-valued weights $W$.

\subsection{Differentiable NAS}

\method adopts a differentiable architecture search flow~\cite{darts, pdarts, pcdarts}. In differentiable NAS, a supernet is constructed such that all possible architectures are sub-architectures of this supernet. Then, architectural choices parameterized by architectural parameters $\alpha$ are optimized following the gradients of the validation loss $L_{val}$. The bilevel optimization problem can be written as
\begin{equation}
\begin{split}
&\min_{\alpha}{L_{val}(w^*(\alpha), \alpha)}\\
\mbox{s.t. } &w^*(\alpha) = \argmin_{w} L_{train}(w, \alpha)
\end{split}
\end{equation}

After the search, one needs to derive a discrete architecture using the relaxed architecture parameter $\alpha$. In the original differentiable NAS method~\cite{darts}, for the normal and reduction cell type, the operation with the maximum $\alpha$ (except for the ``none'' operation) is chosen on each edge. Then the normal and reduction cells are stacked to construct the final model.
Studies~\cite{understanding} have shown that the derive process introduces a large discrepancy. \method also focuses on how to bridge the search-derive gap.
 


\section{BARS Framework}
\label{sec:method}

As we have analyzed before, the unsatisfying performance of BNNs can be attributed to the information bottlenecks that are related to both the macro and micro-level architectural design. 
Due to the high dimension of the architectural design, it is hard to analyze the architectural bottleneck and design suitable BNN architecture manually,  BARS seeks to solve this issue in an automatic manner with differentiable NAS. BARS designs a macro search space (Sec.~\ref{sec:method-macro-ss}) with a learnable cell layout to strike a balance between performance and complexity, and a micro-level search space (Sec.~\ref{sec:method-micro-ss}) tailored for maximizing the information flow. Aside from that, we employ a few search strategies (Sec~\ref{sec:method-strategy}) to address the ``collapse'' problem of differentiable NAS and facilitate a stable search process.

\subsection{Macro-level: Search for Network Depth/Width}
\label{sec:method-macro-ss}

\begin{figure}[t]
    \centering
	\includegraphics[width=\linewidth]{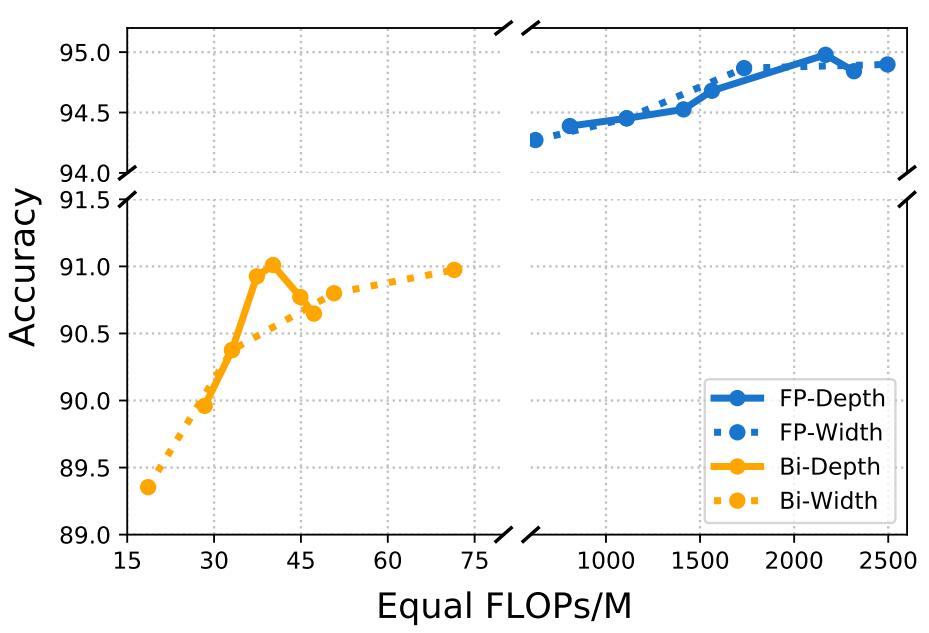}
    \caption{\textbf{Performances of (binary) ResNet18 variants with different complexity.} The original ResNet18 architecture are scaled by uniform expansion in the channel dimension (width), or stacking more layers (depth).}  
    \label{fig:macro-motivation}
\end{figure}

Fig.~\ref{fig:macro-motivation} shows how width/depth configurations affect the performances of XNOR/FP ResNets.
 In general, BNNs are more sensitive to changes in width and depth. And we can see that expanding the operation width can alleviate the bottlenecks of binarized operations and brings consistent improvements. As the width goes up, the performance gain would gradually vanish while still increasing the model complexity.
  In contrast, increasing the depth does not always bring improvements to BNNs, which is intuitive since adding more processing cannot recover the information once the information is already lost in previous bottlenecks.
Due to these distinct preferences of BNNs, directly adopting layouts of CNNs as in~\cite{bnas,bnas-width} might lead to suboptimal designs. And we propose to directly search for the ``sweet spot'' of width and depth to trade-off performance and complexity for BNNs. We extend the original micro-level DARTS framework to the macro-level by designing a macro depth \& width search space that could be unified in the DARTS framework. 


\noindent\textbf{Width Search} Expanding the width of BNNs can bring consistent improvements~\cite{Bethge2018TrainingCB}, and different parts of the network have different sensitivity to the width choices~\cite{hawq}. Previous NAS for BNN studies~\cite{bnas, bats} neglect this aspect in the search process and use post-search uniform width expansion, which would lead to suboptimal results (See Fig.~\ref{fig:macro-ablation}). In contrast, \method seeks to directly search for the proper width configuration to better balance the model complexity and performance.


Specifically, for all cells in one stage, we use a width architecture parameter $\alpha_{width}$ to denote the probability logits of selecting different width choices (e.g. $[0.25, 0.5, 0.75, 1.0]$ in our experiments). Specifically,  during the forward process, we sample a relaxed width choice $m \in \mathcal{R}^4$ from the distribution parameterized by $\alpha_{width}$ with Gumbel-Softmax sampling, and multiply the weighted mask to all the feature maps in that stage. Denoting the full-width output feature map as $y'$, the relaxed output feature map $y$ that takes the width choice into consideration is calculated as



\begin{equation}
  \begin{split}
    {\bf m} =& \mbox{Gumbel-Softmax}(\alpha_{width})\\
    y = & (\sum_{i} {\bf m_i} M_i) \odot y'\\
  \end{split}
  \label{equ:width}
\end{equation}
where $\odot$ denotes the element-wise multiplication, and $M_i$ is the mask corresponding to the $i$-th width choice in $[0.25, 0.5, 0.75, 1.0]$. For example, if $i=0$ (the relative width choice is 0.25), the first quarter of the elements in $M_i$ are $1$s and the other elements are $0$s.

\noindent\textbf{Depth Search}
(Cell-1, Cell-2, Cell-3) in one stage, the probabilities of choosing different depth $\in\{0,1,2,3\}$ can be calculated as the softmax of the four-dimensional depth parameter $\alpha_{path}=[\alpha_{p0}, \alpha_{p1}, \alpha_{p2}, \alpha_{p3}]$: $P(\mbox{depth}=i) = \frac{\exp(\alpha_{pi})}{\sum_{j=0}^3\exp(\alpha_{pj})}$. Denoting the input feature map of this stage as $y_0$, and the output feature map of each cell as $y_1, \cdots, y_3$, the aggregated feature map $y_{aggr}$ is calculated as

\begin{equation}
  \begin{split}
    {\bf m} = &       \mbox{Gumbel-Softmax}(\alpha_{path})\\
    y_{aggr} &= \sum_{i} {\bf m_i} \times y_{i}.
    \end{split}
    \label{equ:depth}
\end{equation}


\noindent\textbf{Complexity Regularization} Besides the performance, the model complexity is also largely influenced by the width/depth macro-level search decisions. Thus, we use a search objective that considers the complexity (FLOPs) to properly balance the performance and complexity. 

\begin{equation}
    \begin{aligned}
        L &= L_0\times\left[\frac{FLOPs(\alpha)}{F}\right]^\theta\\
        \theta &= 
        \begin{cases}
           \gamma,\quad\mbox{if } \xspace FLOPs(\alpha) \geq F  &   \\
           \mu,\quad\text{otherwise}
        \end{cases}
    \end{aligned}
\end{equation}
where $L_0$ is the original loss and $F$ is the FLOPs budget, and $FLOPs(\alpha)$ is the current estimation of FLOPs. $\gamma$ and $\theta$ are hyperparameters and $\gamma$ is set to 0 in our experiments. 

\begin{table}[t]
    \centering
    \begin{tabular}{c|c}
    \toprule
    Method & Accuracy \\
    \midrule
        FP downsample  &  90.6\%  (-\%) \\
        with op shortcut & 91.4 (+0.8\%)\\
		binarized downsample & 89.7\% (-0.8\%)  \\
        improved binarized downsample & 89.9\% (-0.6\%)  \\
	\midrule
		XNOR-Res34 ($\sim$1.5x complexity) & 91.5\% (+0.9\%) \\
    \bottomrule
    \end{tabular}
    \vspace{6pt}
    \caption{\textbf{Performance with micro modifications applied on Binary ResNet18.} Binarizing the downsampling layer causes noticable accuracy drop and improved binary downsampling could mitigate the loss. Adding op-wise shortcut could improve the performance without extra overhead, and achieve competitive performances with a larger model.}
    \label{tab:micro-motivation}
\end{table}

\subsection{Micro-level: Search for Cell Topology}
\label{sec:method-micro-ss}

As mentioned above, identifying and eliminating the information bottleneck is vital for improving the accuracy of BNNs~\cite{Bethge2018TrainingCB}. Since it is difficult to assess and identify all information bottlenecks manually, we design the micro-level search space such that the NAS process can automatically discover topologies to avoid creating bottlenecks and maintain a proper information flow.
As illustrated in Fig.~\ref{fig:main-flow}, our micro-level search space contains five nodes, and each node can choose to connect from an arbitrary number of previous nodes. For each edge, there are 3 possible operation primitives: binary convolution 3$\times$3, shortcut, and none. We will show in Sec.~\ref{sec:cell-ablation} that our micro-level topology search automatically discovers cells with a strong information flow. 

Besides the previous search space design, the cell template and operation primitives should also be modified to eliminate information bottlenecks. 


 


\begin{table*}[t]
  \vspace{-10pt}
  \centering
  \label{tab:results}
  \begin{tabular}{c|c|c|c|c|c|c|c}
    \toprule
    {Dataset} & {Method} & {FLOPs} & {BiOps} &  {Equivalent Ops} & {Params} & {Fully-Binarized} & Acc.  \\
    \midrule
    \multirow{10}{*}{CIFAR-10}& NiN (XNOR) & $\sim$9M & $\sim$200M & 12.13M & -  & \checkmark & 86.28\% \\
    &ResNet-18 (XNOR) & 16M & 547M & 25.55M & 11.17M & \checkmark & 90.55\% \\
    &ResNet-18 (Bireal) & 11M & 561M & 24.77M & 11.34M & FP Downsample & 91.23\% \\
    &ResNet-34 (XNOR) & 27M & 1151M & 44.98M & 21.81M & FP Downsample & 91.49\% \\
    &WRN-40 (XNOR)  & $\sim$27M & $\sim$1500M & 50.44M &  - & FP Downsample & 91.58\% \\
    & ResNet-18  (IRNet) & $\sim$27M & $\sim$500M & 34.81M & - & FP Downsample & 91.5\% \\
    &BNAS (XNOR)  & 100M & 1393M & 121.76M & 5.57M & FP Cell Shortcut & 92.7\% \\
    \cmidrule{2-8}
    & BARS-A  & 2M & 513M & 10.02M & 2.77M & \checkmark & \textbf{91.25\%} \\
    & BARS-B  & 2M & 1048M & 18.37M & 6.07M & \checkmark & \textbf{92.98\%}  \\
    & BARS-C  & 3M & 1778M & 32.27M & 10.76M &  \checkmark & \textbf{93.43\%} \\
    \midrule
    \multirow{8}{*}{ImageNet} & ResNet-18 (ABC) & $\sim$100M & $\sim$2000M & 131.25M & - & \checkmark & 42.7\% \\
    & ResNet-18 (XNOR) & 138M & 1850 & 188.89M & 12.54M & \checkmark & 48.3\% \\
    & ResNet-18 (XNOR) & 167M & 1778M & 194.79M & 12.80M & FP Downsample & 53.1\% \\
    & BiDenseNet (XNOR) & - & - & -  & 13.56M & FP Downsample & 52.7\% \\
    & ResNet-18 (Bireal) & $\sim$160M & $\sim$2000M & 191.25M & - & FP Downsample & 56.4\% \\
     & ResNet-18 (PCNN) & $\sim$160M & $\sim$2000M & 191.25M & - & FP Downsample & 57.3\% \\     
     & MoBiNet(XNOR) & - & - & - & 8.47M & FP Downsample & 53.4\% \\
     & BNAS (XNOR) & 195M & 3137M & 244.0M & 28.41M & FP Cell Shortcut & 57.6\% \\
     \cmidrule{2-8}
     & BARS-D & 129M & 998M & 129.60M & 9.01M & \checkmark & \textbf{54.6\%} \\
     & BARS-E & 161M & 1424M & 183.25M & 14.04M & \checkmark & \textbf{56.2\%} \\
     & BARS-F & 254M & 2594M & 293.53M & 19.29M & \checkmark & \textbf{60.3\%} \\
     \bottomrule
  \end{tabular}
  \vspace{10pt}
    \caption{\textbf{Performance and complexity comparison of \method-discovered architectures and baselines on CIFAR-10 and ImageNet.} \method-discovered architectures outperform baseline architectures with much lower resource consumption. ``Equivalent Ops'' are calculated as $FLOPs+1/64*BiOPs$~\cite{Bethge2018TrainingCB}. ``Fully-Binarized'' means all network components except for the first and last layer remain binary. Unlike recent studies~\cite{bireal,reactnet,bnas} on binary architectures use full-precision downsampling operations, \method-discovered architectures only use binarized operations in the major architecture backbone, which is beneficial for hardware acceleration.  }
\end{table*}

\noindent\textbf{Strengthen the information flow} Many recent studies on BNNs~\cite{Bethge2018TrainingCB, bireal} and our experimental results in Tab.~\ref{tab:micro-motivation} show that adding an extra shortcut for binary convolution improves the performance with little overhead. Therefore, we add operational-level shortcuts and cell-level shortcuts to strengthen the information flow. 


\noindent\textbf{Eliminating the information Bottleneck} 
As shown in Tab.~\ref{tab:micro-motivation}, binarizing the downsampling layer causes the performance to drop sharply. There are two reasons why the downsampling layer is the bottleneck: 1) The misalignment of input/output channel size makes it difficult to add op-wise shortcuts. 2) It involves spatial dimension reduction which causes a large information loss.
In \method, we design the downsampling operation to concatenate the outputs of two strided convolutions with shortcuts\footnote{We use 2$\times$2 AvgPool as the shortcut with spatial dim. reduction.} and spatially staggered input. In such way, the op-level shortcut could be added to the two binarized convolutions. 

It is worthy to note that, unlike many previous studies~\cite{bireal,xnornet} that use FP downsampling layers, we mitigate this bottleneck issue in a fully-binarized manner. And we demonstrate its effect in Tab.~\ref{tab:micro-ablation}

\subsection{Search Strategy}
\label{sec:method-strategy}


Differentiable NAS~\cite{darts} has been challenged due to it being prone to degenerated architectures containing too many shortcuts, which is known as the ``collapse'' problem. Parameterized operations (e.g. Conv) are usually under-trained~\cite{understanding,tfnas}, and the search process will favor parameter-free operations.  
In BNN, this problem is further exacerbated since binary convolutions are even harder to train. 
This section describes several key techniques we use to alleviate the ``collapse'' problem. 
We briefly introduce them here and further analyze their effects in Sec.~\ref{sec:stabilize}.


\noindent\textbf{Gumbel-Softmax Sampling} 
We use Gumbel sampling with proper temperature scheduling for all architecture decisions. For all architectural parameters (depth $\alpha_{path}$, width $\alpha_{width}$, operation type $\alpha$), we sample a relaxed architecture $m$ from the corresponding multinomial architectural distribution $\mbox{Multinomial}(m|\alpha)$ with the Gumbel-Softmax technique~\cite{jang2016categorical}. Denoting the number of choices as $D$, the logits of the Multinomial distribution as $\alpha$, each dimension $m_i$ in the relaxed architecture decision $m \in [0,1]^D$ can be represented as
\begin{equation}
  \begin{split}
    m_i = \frac{\exp{((\alpha_i + g_i)/\tau)}}{\sum_{j=1}^D \exp( (\alpha_j + g_j)/\tau)}\quad \mbox{for } i=1,\cdots D,
  \end{split}
  \label{equ:gumbel}
\end{equation}
where $g_i$s are standard Gumbel-distributed random variables. We emphasize that using Gumbel-Softmax sampling with a proper temperature schedule is important.
In the early search stage, it remains high 
which drives architecture distributions to be more uniform to encourage search exploration and avoid collapsing. In the later stage, it gradually anneals to zero to drive $\alpha$ towards confident one-hot choices. Thus the discrepancy between searching and deriving~\cite{stablizing} is reduced.

\noindent\textbf{Entropy Regularization and Supernet Warm-up} To address the ``collapse'' issue caused by insufficient optimization of parameterized operations, we conduct warm-up training of the supernet weights for several epochs. We also impose entropy regularization on the architecture distribution to encourage exploration in the early search stage and exploitation in the late search stage.

\begin{equation}
    \begin{split}
        L_{ent} =  \lambda_{ent} ( - \sum_{i}{\alpha_i \log(\alpha_i) ) },
    \end{split}
\end{equation}
where $\lambda_{ent}$ is a hyperparameter that follows an increasing schedule from negative to positive. Its scheduling plays a similar role as the temperature in gumbel sampling.

\section{Experiments and Analysis}
\label{sec:exp}

\subsection{Experiment Settings}
\label{sec:exp-setting}

We run \method on CIFAR-10 and ImageNet datasets with different FLOPs target, and acquire a series of models of different sizes (\method-A/B/C on CIFAR, \method-D/E/F on ImageNet). Detailed experimental settings can be found in the appendix. Note that unlike previous studies~\cite{bnas, bats} that transfer architectures found on CIFAR-10 to ImageNet, we directly apply search on the 100-class subset of ImageNet.

We conduct experiments on CIFAR-10 and ImageNet. For searching on both datasets, we construct a 14-cell super network organized into 3 stages. The cells in each stage share the same micro-level topology, and so do all the reduction cells. The base channel number $C_i$ is 48. We choose 4 available width choices $r \times C_i, r\in\{0.25, 0.5, 0.75, 1\}$, and 3 candidate operations $[binary\_conv\_3x3, shortcut, none]$. 
Within each cell, the preprocess layer is a binary 1x1 conv with no shortcut. The cell-wise shortcut is an identity operation for normal cells and a strided binary 3x3 conv for reduction cells.

\begin{figure}
    \centering
    \includegraphics[width=\linewidth]{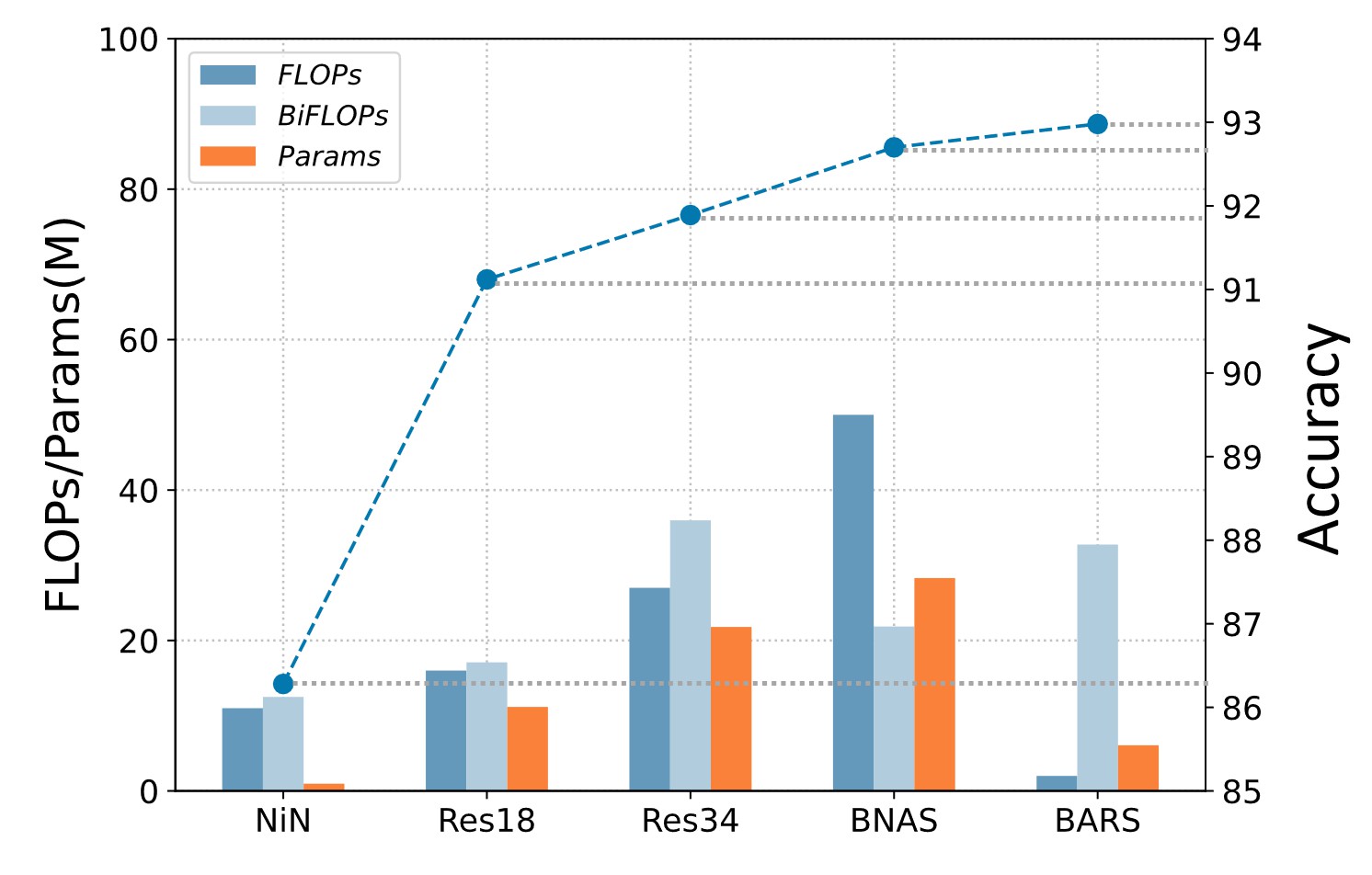}
    \caption{\textbf{Comparison of \method derived architectures with baseline methods in terms of FLOPs (floating-point ops), BiOPs (binary ops), and Params.} Note that for BiOPs, following previous studies~\cite{Bethge2018TrainingCB}, we divide it by 64 to acquire its relative values. \method finds architectures with better performance with less equivalent OPs/params and significantly fewer floating-point operations. }
    \label{fig:bar}
    \vspace{-10pt}
\end{figure}

The search lasts for 50 epochs, and a batch size of 64 is used.
Supernet weights $w$ is trained with Adam optimizer, whose learning rate is set to 3e-4 initially and decayed to 0 in 50 epochs following a cosine schedule. 
After 5 epochs of warm-up training of supernet weights, we begin to update $\alpha$. The architectural parameters $\alpha$ (including $\alpha_{micro}, \alpha_{path}, \alpha_{width}$) are updated using Adam optimizer with learning rate 3e-4 and weight decay 1e-3. The Gumbel temperature is set to $1$ at first and multiplied by $0.9$ on every epoch. The entropy regularization coefficient $\lambda_{ent}$ follows an increasing schedule: it starts at -0.01, and 0.001 is added on every epoch. 

As for deriving, we sample 8 candidate architectures from the architecture distribution 
$\alpha$ after the search. The one with minimum validation loss is chosen.
Different from the origin DARTS, we do not need to exclude the ``none'' operation when deriving. We argue that it is important since binary convs evolve huge information loss, thus ``none'' operation might be the best choice on some edges in BNN~\cite{bnas}.
On CIFAR-10, we train the derived architecture for 200 epochs with batch size 256. Adam optimizer with a weight decay of $0$ is used, and the learning rate is set to $2e-3$ at first and decayed to 0 following a cosine schedule. Cutout augmentation and auxiliary towers with weight 0.4 are applied following previous studies~\cite{darts}. On ImageNet, the architectures are trained for 100 epochs, and no cutout is used. We also use Adam optimizer with no weight decay. The learning rate has the initial value of $1.e-3$ and cosine annealed to 0 for a batch size of $256$.

\begin{figure}[t]
    \centering
    \includegraphics[width=0.95\linewidth]{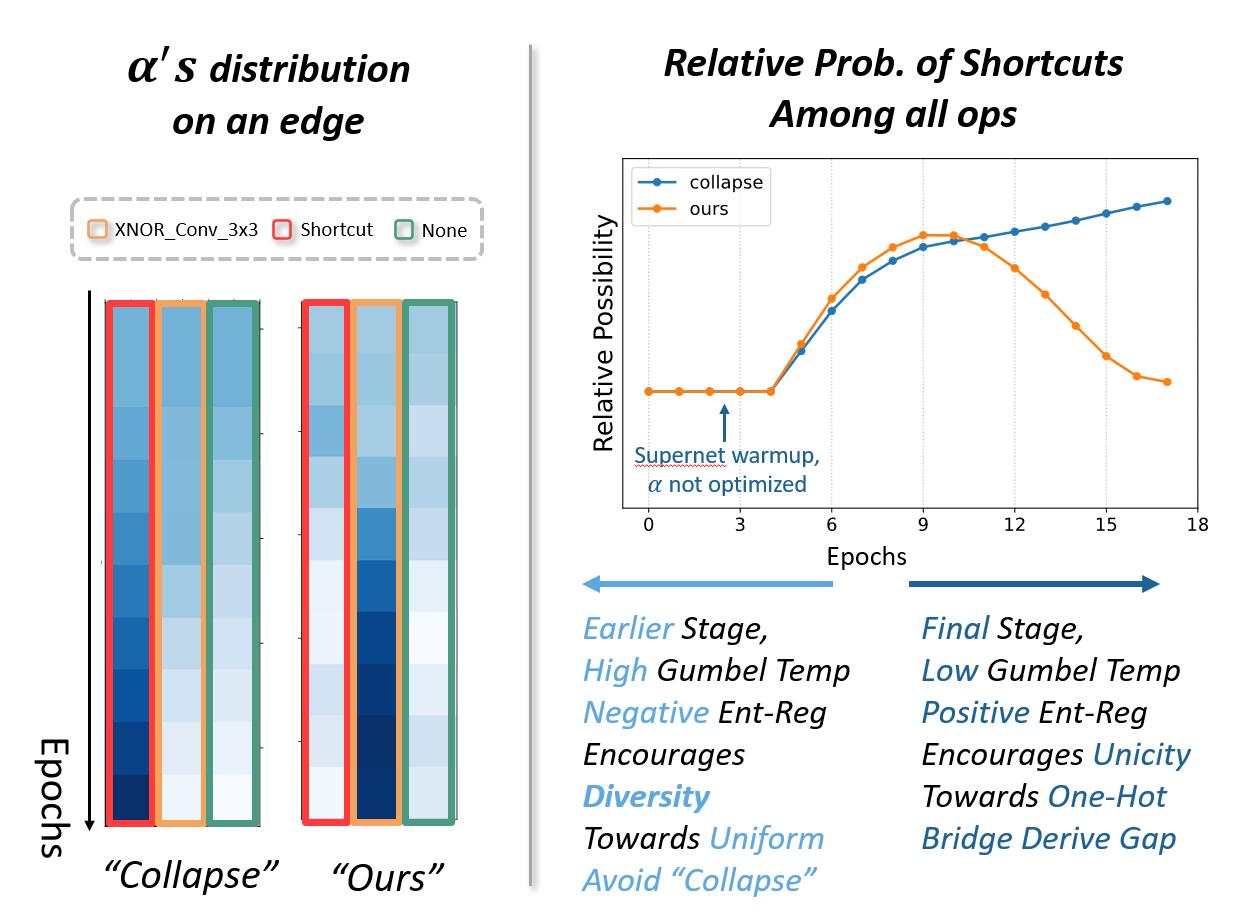}
    \caption{\textbf{Evolution of $\alpha$'s distribution and the relative possibility of ``shortcut'' in the search process} Left: Comparison of $\alpha$'s distribution change for ``collapsed search'' and \method's stabilized search. Entropy regularization and Gumbel-Softmax sampling prevent ``collapse'' in search and bridge the gap in deriving. Right: ``shortcut'' operation's average probability during the search. \method prevents the search from collapsing rapidly.} 
    \label{fig:alphas}
\end{figure}

\subsection{Results on CIFAR-10 and ImageNet}
Tab.~\ref{tab:results} and Fig.~\ref{fig:bar} show the comparison of \method-discovered architectures and the baseline ones. 
We can see that \method discovers architectures with superior performance and efficiency. Note that in order to demonstrate the performance gain brought by the architectural design, all our models are trained from scratch with XNORNet~\cite{xnornet} binarization scheme.
Neither additional tricks~\cite{irnet}, nor full-precision pre-training models~\cite{bireal,PCNN} are used.
Moreover, we emphasize that \method binarizes all operations in the major architecture (except the stem and the final layer), whereas previous studies use full-precision downsampling layers to maintain acceptable performances (e.g. the accuracy of ResNet-18 on ImageNet dropped from $53.1\%$ to $48.3\%$ if no FP downsample is used).
On CIFAR-10, \method-B achieves higher accuracy ($1.5\%$) than the hand-crafted binary model with $2/3$ binary operations (BiOps), much less ($1/10$) floating-point operations (FLOPs) and parameters.
On ImageNet, \method-D outperforms the ``fully-binarized'' ResNet-18 by a 6\% with notably less resource consumption, it also outperforms many hand-crafted BNN models while binarizing the downsampling layer.


\textbf{
\begin{figure}[t]
    \centering
    \includegraphics[width=\linewidth]{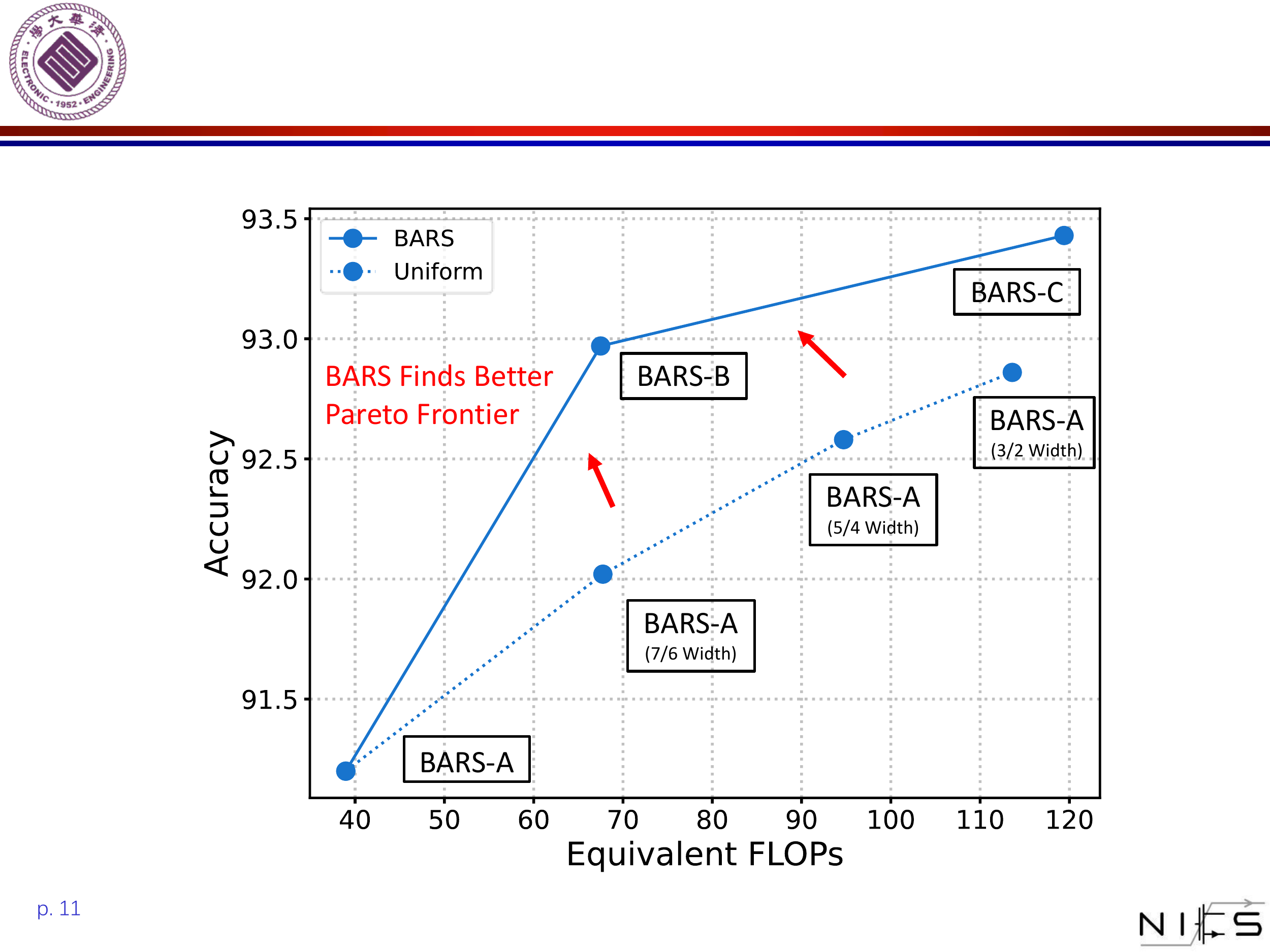}
    \caption{\textbf{Comparison of derived models of different complexity (FLOPs) w/o \method's complexity search.} Upper: \method search with different complexity regularization. Lower: search for a compact model and uniformly expand the network width. \method finds the better pareto frontier.          
    }
    \label{fig:macro-ablation}
\end{figure}
}

\subsection{Effects of Stabilizing the Searching}
\label{sec:stabilize} 

The ``collapse-to-shortcuts'' problem is a widely known issue for differentiable NAS methods, as shown in the left part in Fig.~\ref{fig:alphas}.
As discussed in Sec.~\ref{sec:method-strategy}, we apply warm-up training to prevent the search from collapsing in the very early stages.
Then, entropy regularization and Gumbel sampling with proper hyperparameter scheduling are employed to encourage exploration in the early stages and confident decisions in the late stages. 
Fig.~\ref{fig:alphas} shows that in the early stage of searching, the distribution of $\alpha$ is relatively uniform and the relative ranking of different operations keeps changing. When reaching the end of the search, the Gumbel temperature is close to zero, making the sampling close to one-hot sampling. Also, entropy regularization encourages architecture distribution to be more confident. This reduces the derive discrepancy. 

\begin{table}[t]
    \centering
    \begin{tabular}{c|c}
    \toprule
    Method & Accuracy \\
    \midrule
        BARS-B & 93.0\% \\
        BARS-B (without op shortcut) & 92.8\% (-0.2\%) \\
        BARS-B (without improved ds.) & 92.6\% (-0.3\%) \\
        \midrule
        Sampled arch. & 92.9\% \\
        Sampled arch. (without op shortcut) & 89.8\% (-3.1\%)  \\
        Sampled arch. (without improved ds.) & 91.9\% (-1.0\%) \\
    \bottomrule
    \end{tabular}
    \caption{\textbf{The ablation study of micro-level modifications for BARS-B and sampled architecture.} Searched BARS-B model finds that each shortcuts should coordinate with binary convs, thus the performance does not drop much without the op-level shortcut. }
    \label{tab:micro-ablation}
\end{table}

\subsection{Effects of the Search Space Design}
\label{sec:ss-ablation}

We conduct several experiments to verify the effectiveness of the design choices of our search space. As could be seen from Fig.~\ref{fig:macro-ablation}, the macro-level joint search of width and depth strikes a better balance between performance and complexity. The discovered BARS-B model achieves much higher accuracy than uniformly expanding the width of the smaller BARS-A model. 

For the micro-level design, we have shown that additional shortcut and improved binary downsample can bring performance gain for XNOR-ResNet18 in Tab.~\ref{tab:micro-ablation}. We conduct similar ablation studies on BARS-B and a random sampled architecture from our search space in Tab.~\ref{tab:micro-motivation}. And we can witness a noticeable accuracy degradation when removing our modifications.

\subsection{Discovered Cell}
\label{sec:cell-ablation}

The \method-discovered cells on CIFAR-10 are shown in Fig.~\ref{fig:cells}. We can see that the cells in earlier stages and the reduction cells contain more convolutions. Conversely, the latter cells are dominated by shortcuts. 
The micro-level topology search also discovers interesting connection patterns: the shortcuts coordinate with binary convs to strengthen the information flow (e.g. shortcut 1-4 and shortcut 2-4 around conv 1-2 in the Normal-1 cell). Thanks to this highly skip-connected pattern, we can see from Tab.~\ref{tab:micro-ablation} that removing the op-level shortcut causes much less performance degradation for the \method-discovered cell than for randomly sampled architecture. Also, instead of transferring the architecture discovered on CIFAR-10 to ImageNet, \method conducts a direct search on the 100-class subset of ImageNet, and discovers distinct cells with more convolutions (see the appendix for more details). The different cell preferences might result from that more parameterized operations are needed for enough representational ability on the larger Imagenet dataset.


\begin{figure}[bt]
    \centering
    \includegraphics[width=\linewidth]{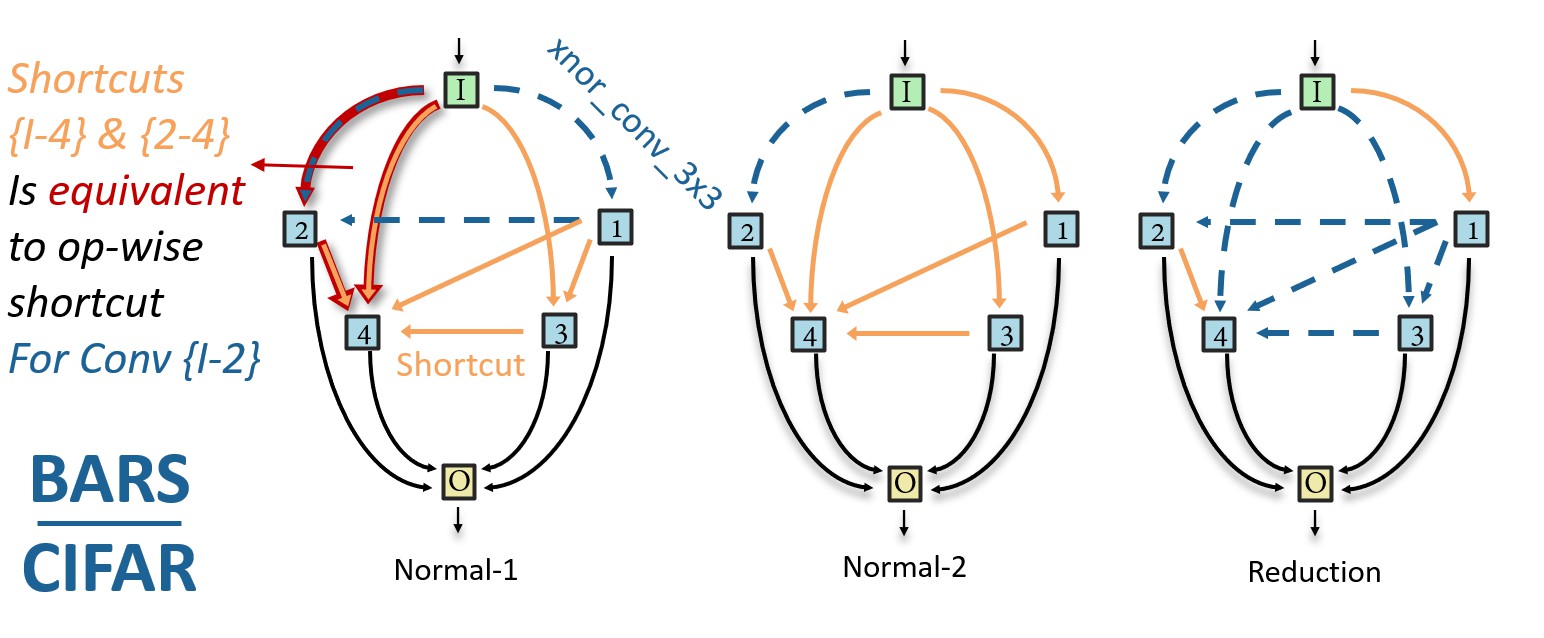}
    \caption{
    \textbf{BARS discovered cells on CIFAR. } More convs are found in the reduction cell and shallower part of the network. It also learns that shortcuts should coordinate with binary convs. }  
    \label{fig:cells}
\end{figure}


\section{Conclusion}
\label{sec:conclusion}

To better explore BNN architectures that are both accurate and efficient, \method proposes
to use a joint search of the macro layout and the micro topology to address the information bottleneck problem in BNN. 
The binary architectures discovered by \method outperform baseline architectures with significantly less resource consumption.

\clearpage

{\small
\bibliographystyle{ieee_fullname}
\bibliography{egbib}
}

\clearpage

\begin{appendices}
\section{Experiment Settings}
We conduct experiments on CIFAR-10 and ImageNet. For searching on both datasets, we construct a 14-cell super network (a.k.a, ``supernet'') organized in 3 stages. The cells in one stage share the same micro-level topology, and all reduction cells share one micro-level topology. The base channel numbers of the cells in the $i$-th stage $C_i$ is 48, 96, and 192 for $i=1,2,3$, respectively. Available width choices of the cells in the $i$-th stage are $r \times C_i, r\in\{0.25, 0.5, 0.75, 1\}$.
Within each cell, the preprocess layer is a binary 3x3 convolution with no shortcut. The cell-wise shortcut is an identity operation for normal cells and a strided binary 3x3 convolution for reduction cells.

The search lasts for 50 epochs, and a batch size of 64 is used.
Following the standard practice of differentiable NAS~\cite{darts}, half of the training dataset is used to update the supernet weights $w$, and the other half is used as the validation dataset to update the architecture parameters $\alpha$. For ImageNet, we randomly sample 100 classes of data for searching the architecture, and the whole dataset is used for training. Supernet weights $w$ is trained with Adam optimizer, whose learning rate is set to 3e-4 initially and decayed to 0 in 50 epochs following a cosine schedule. The weight decay is set to 0 as for BNN training~\cite{bireal}. After 5 epochs of warm-up training of supernet weights, we begin to update $\alpha$. The architectural parameters $\alpha$ (including $\alpha_{micro}, \alpha_{path}, \alpha_{width}$) are updated using Adam optimizer with learning rate 3e-4 and weight decay 1e-3. The Gumbel temperature is set to $1$ at first and multiplied by $0.9$ on every epoch. The entropy regularization coefficient $\lambda_{ent}$ follows an increasing schedule: it starts at -0.01, and 0.001 is added on every epoch. The $\beta$ parameter in the capacity regularization is set to $0.2$.

As for the deriving, after the search, we sample 8 candidate architectures from the architecture distribution parametrized by $\alpha$. Then, the one with minimum valid loss is chosen as the derived architecture. On CIFAR-10, we train the derived architecture for 200 epochs with batch size 256. Adam optimizer with a weight decay of $0$ is used, and the learning rate is set to $2e-3$ at first and decayed to 0 following a cosine schedule. Cutout augmentation and auxiliary towers with weight 0.4 are applied following previous studies~\cite{darts}. On ImageNet, the architectures are trained for 100 epochs, and no cutout is used. We also use Adam optimizer with no weight decay. The learning rate has the initial value of $1.e-3$ and cosine annealed to 0 for a batch size of $256$.

\begin{figure*}[t]
    \centering
    \includegraphics[width=0.98\linewidth]{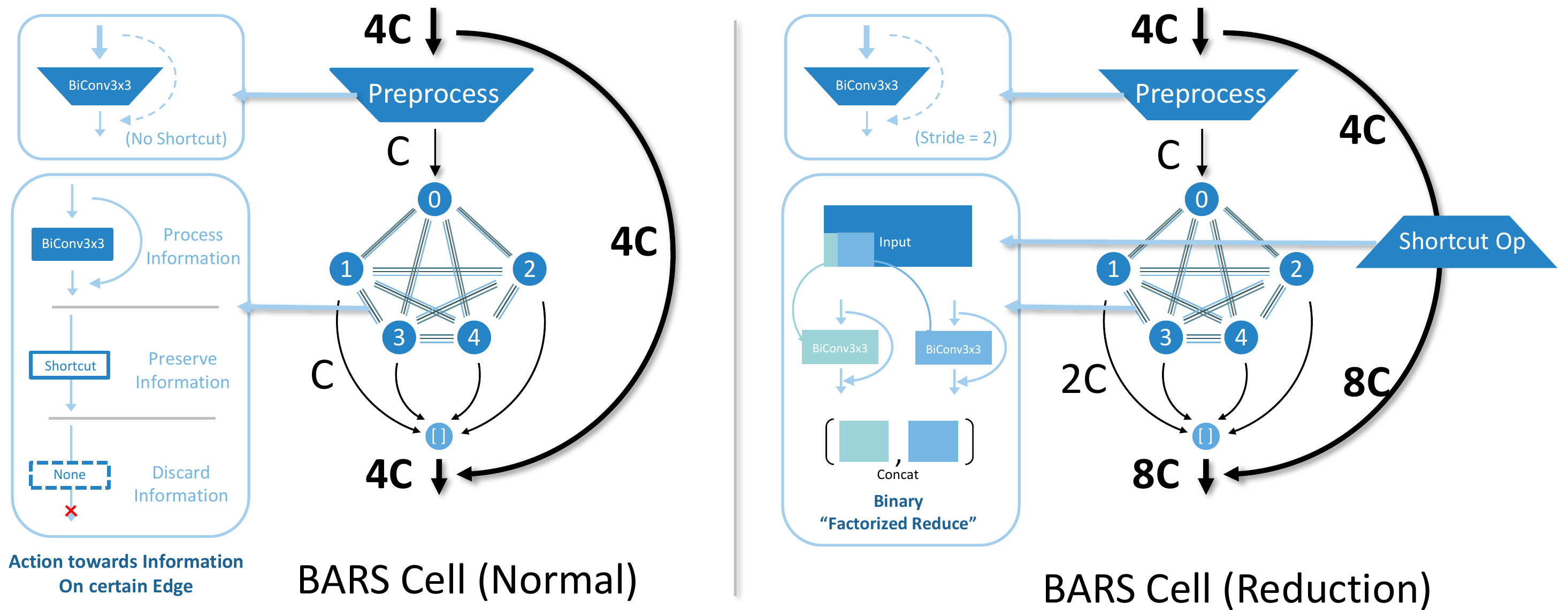}
    \caption{\textbf{Illustration of cell template and operations}}
    \label{fig:ops}
\end{figure*}

\section{Candidate Operations}


In BARS, we choose 3 candidate operations: None, Shortcut and Binary Conv 3x3. 

As discussed in the paper, previous studies~\cite{bireal} have witnessed the importance of shortcut. In order to strengthen the information flow, we add the shortcut from the input binary convolution 3x3 to its output. In normal cells, the shortcut is simply the Identity transform since the input and output are of the same size. In reduction cells, when using plain binary convolution with a stride of 2, the output has spatial size reduction and 2x width expansion. In BARS, inspired from the ``factorized reduce'' layer in ResNet, we propose the binary version of the ``factorized reduce'' as illustrated in Fig.~\ref{fig:ops}. It consists of two strided convolutions with shortcut and spatially staggered input. Their outputs are concatenated in the width dimension as the final output.

\section{Detailed Comparison of Searched Cells}

Unlike previous studies that transfer the architecture discovered on CIFAR-10 to ImageNet, we directly conduct the search on a 100-class subset of the ImageNet dataset, since the dataset distributions of the two datasets are notably different. As could be witnessed in Fig.~\ref{fig:cells}, the architectural preferences on CIFAR-10 and ImageNet are quite different. The model discovered on CIFAR has relatively fewer convolutions. On the contrary, the one on ImageNet is more ``dense'' and convolutions are dominant in most of the cells.  

\begin{figure*}[h]
    \centering
    \includegraphics[width=0.95\linewidth]{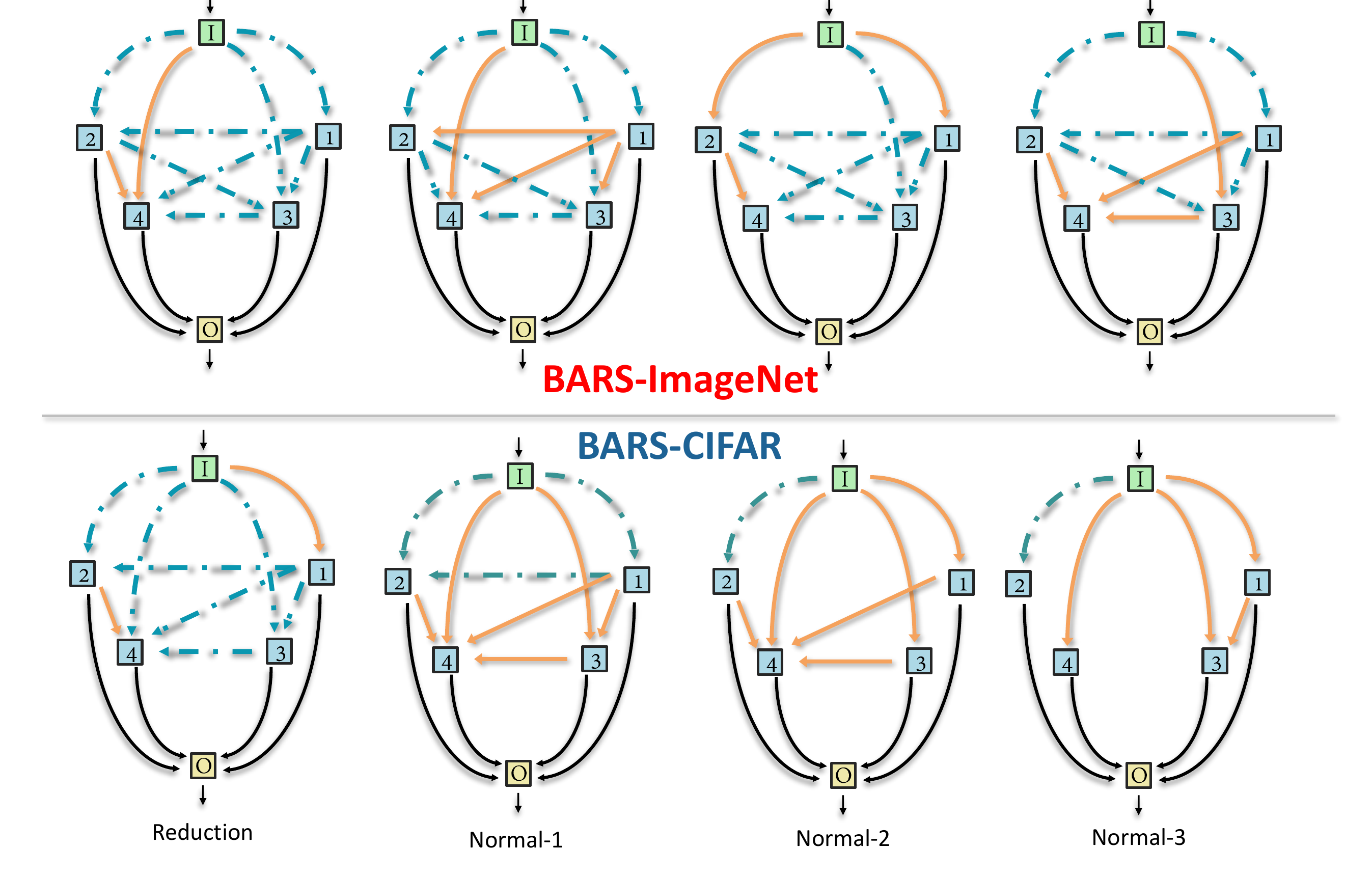}
    \caption{\textbf{Comparison of searched cell on CIFAR and ImageNet}}
    \label{fig:cells}
\end{figure*}
\end{appendices}

\end{document}